\documentclass{article}
\usepackage[utf8]{inputenc} % allow utf-8 input
\usepackage[T1]{fontenc}    % use 8-bit T1 fonts
\usepackage{hyperref}       % hyperlinks
\usepackage{url}            % simple URL typesetting
\usepackage{booktabs}       % professional-quality tables
\usepackage{amsfonts}       % blackboard math symbols
\usepackage{nicefrac}       % compact symbols for 1/2, etc.
\usepackage{microtype} 
\usepackage{bm}
\usepackage[pdftex]{graphicx}
\usepackage{siunitx}
\usepackage{array, amsmath}
\usepackage{authblk}

\begin{document}

\title{Learning second order coupled differential equations that are subject to non-conservative forces}
\author[1]{R. A. M\"uller}
\author[1]{J. Laflamme-Janssen}
\author[1]{J. Camacaro}
\author[1]{C. Bessega}

\affil[1]{StradigiAI}
\date{Decemebr 19, 2019}   
\maketitle

\begin{abstract}
In this article we address the question whether it is possible to learn the differential equations describing the physical properties of a dynamical system, subject to non-conservative forces, from observations of its real-space trajectory(-ies) only. We introduce a network that incorporates a difference approximation for the second order derivative in terms of residual connections between convolutional blocks, whose shared weights represent the coefficients of a second order ordinary differential equation. We further combine this solver-like architecture with a convolutional network, capable of learning the relation between trajectories of coupled oscillators and therefore allows us to make a stable forecast even if the system is only partially observed. We optimize this map together with the solver network, while sharing their weights, to form a powerful framework capable of learning the complex physical properties of a dissipative dynamical system.

\end{abstract}

\section{Introduction and Review}
Dynamical systems find manifold applications to processes in everyday life and allow insights into many areas not only of physics, but also mathematics, or theoretical biology. A dynamic system can be described by a mathematical model of a time-dependent process whose further course depends only on its initial state, but not on the choice of the starting point in time. In this work we focus on dynamical systems that can be described by ordinary differential equations (ODEs).\\

If the system is only subject to conservative forces, i.e. forces that can be derived from a scalar potential, higher order ODEs can be written as first order systems by introducing derivatives as new dependent variables of the system.  
An example of such a reduction are Hamilton's equations (Equation \ref{eq:hamilton}). These equations relate the canonical coordinates, position and momentum $ (\boldsymbol {q},\boldsymbol {p}) $, of a system to its total energy,  $\mathcal {H}(\boldsymbol {q},\boldsymbol {p}) = T(\boldsymbol {p})+ V(\boldsymbol {q})$ (for a 1-dim. system with one particle of mass $m$, $T= \boldsymbol {p}/2m $ and $V = V(\boldsymbol {q})$).

\begin{equation}
	\frac {\mathrm {d} \boldsymbol {p}}{\mathrm {d} t}=-\frac {\partial {\mathcal {H}}}{\partial {\boldsymbol {q}}}\quad ,\quad 	\frac {\mathrm {d} {\boldsymbol {q}}}{\mathrm {d} t}=+\frac {\partial {\mathcal {H}}}{\partial {\boldsymbol {p}}}.
	\label{eq:hamilton}
\end{equation}

There have been several recent articles that focus on dynamical systems, only subject to conservative forces, in the context of machine learning \cite{ Lu2018, Long2018, Chen2018, Greydanus2019, Dupont2019}. The fact that such systems are reducible to first order ODEs qualifies them as interesting examples to be studied with residual type networks. It has been first observed by \cite{E2017} that a ResNet block can be understood as a difference equation that approximates a first order ODE.

Residual Neural Networks \cite{He2016} are deep neural networks defined by stacking network blocks combined with a residual connection, adding the input of the block to its output (see left panel of Figure \ref{fig:architecture}).
If we assume a 1-layer convolutional network with $K$ hidden nodes and $\omega_h$, $h=1,...,K$ filter, applied on a 1-dimensional input $(x_t)_{t=0}^{N}$, this becomes:

\begin{equation}
	 x_{t+1, h} = x_t + \text{\emph{NL}} \bigg( \sum_{j=0}^{\infty} \omega_{h ,j}^{l=1}x_{t-j}\bigg),
	 \label{eq:euler0}
\end{equation}
or:
\begin{equation}
	x_{t+1} = x_{t} + F(x_{t}, \Theta_t) 
	\label{eq:euler1}
\end{equation}

with $\Theta_t$ representing the learned network parameters and \emph{NL} the chosen non-linearity. If we have more than one layer in the network 
and $x_l$ represents the hidden state of the $l^{th}$ layer, we can rewrite  Equation \ref{eq:euler1} by introducing a strictly positive parameter $d$, representing the time discretization of a given observable: $d = $ timestep/number of layers.

\begin{equation}
	x_{l+1} = x_l+ d\tilde{F}(x_l, \Theta_l).
	\label{eq:euler2}
\end{equation}

For sufficiently small $d$, the residual block can be interpreted as a forward Euler discretization of a first order differential. In this sense Equation \ref{eq:euler2} builds the link between the network architecture and the dynamics of the system under observation.

The family of problems captured by this formalism is limited to systems that are not subject to dissipation. In this note we are widening the scope and target non-reducible, second-order dynamical systems, i.e. systems that are subject to non-conservative forces as for example friction. We introduce a new network that incorporates the notion of a second-order time derivative, as a function of the systems real space observable and it's time derivative. We then show that we can learn the systems approximate dynamics from its real space trajectory only. We then further ask the question if we still can learn the governing physics of a coupled dynamic system driven by non-conservative forces, if we only partially observe its real-space trajectories?\\ 
From here the paper is structured in the following way: In Section \ref{architecture} we introduce the concept of how residual connections can be used to emulate a second-order finite difference step. In Section \ref{experiments} we put the architecture to the test and Section \ref{exploit} deals with the question if we can further exploit the mathematical relation between real-space trajectories of coupled systems to approximate the differential equations of such a system if observed only partially.

\section{Architecture}\label{architecture}

The base architecture of the network proposed, is inspired by a finite difference solver. We are looking for solutions of a second order (linear) problem of the form:

\begin{equation}
	\ddot{x}+p(t)\dot{x}+q(t) x = r(t)
\end{equation}

To be able to feed our network with real-space trajectory coordinates of a dissipative dynamical system, we need to incorporate the notion of a second order derivative into our architecture that we call \emph{OscillatorNet}. We use the second order central difference approximation, where $\Delta$ represents the time discretization or sampling frequency of a time-series,

\begin{equation}
	\ddot{x}_t \approx  \frac{x_{t+\Delta}-2x_t+x_{t-\Delta}}{\Delta^2} + \mathcal{O}(\Delta^2),
	\label{eq:second_order}
\end{equation}

to retrieve the residual architecture as:

\begin{equation}
	x_{t+\Delta} = x_t+(x_t-x_{t-\Delta}) - F(x_t, \Theta_t) \cdot \Delta^2
	\label{eq:secon_order_arch}
\end{equation}

\begin{figure}[htb!]
	\centering
	\includegraphics[width=1\linewidth]{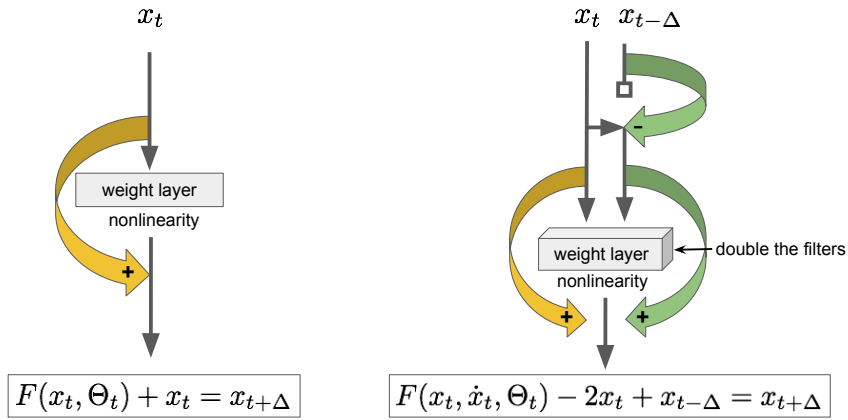}
	\caption{\emph{Left}: Schematic of the ResNet architecture. \emph{Right}: Schematic of the OscillatorNet architecture. The yellow arrow indicates the residual connection, the green arrows indicate the \emph{differential residual}}.
	\label{fig:architecture}
\end{figure}

The right panel in Figure \ref{fig:architecture} shows a schematic of a second order differential block. Additional to the residual connection (yellow arrow) we add a \emph{differential residual} (green arrows), i.e. we subtract in a separate channel $\mathbf{x}_{t-1}$ from $\mathbf{x}_t$ and then add it together with the residual to the network-block output, therefore implementing a second order finite difference step. The differential residual only adds the terms necessary to approximate a second order differential. The network therefore does not have parameters available to learn dissipation (see top panel of Figure \ref{fig:single_oscillator}). By doubling the number of filters we not only learn the inhomogeneous term as a function of the time series variable, i.e. the position (e.g. the force applied by a spring in an elastic pendulum; see lower panel of Figure \ref{eq:setup}) but it also becomes possible, to learn a second inhomogeneous term as a function of the derivative of the time series, i.e. the speed (e.g. the friction force acting on an elastic pendulum). This allows the system to numerically approximate the full differential equation generating the data:

\begin{equation}
	\ddot{x} = f(x,\dot{x}).
	\label{eq:dissipation}
\end{equation} 

\section{Learning ODEs from real-space trajectories}\label{experiments}
\subsection{Damped harmonic oscillator}

\begin{figure}[htb!]
	\centering
	\includegraphics[width=0.5\linewidth]{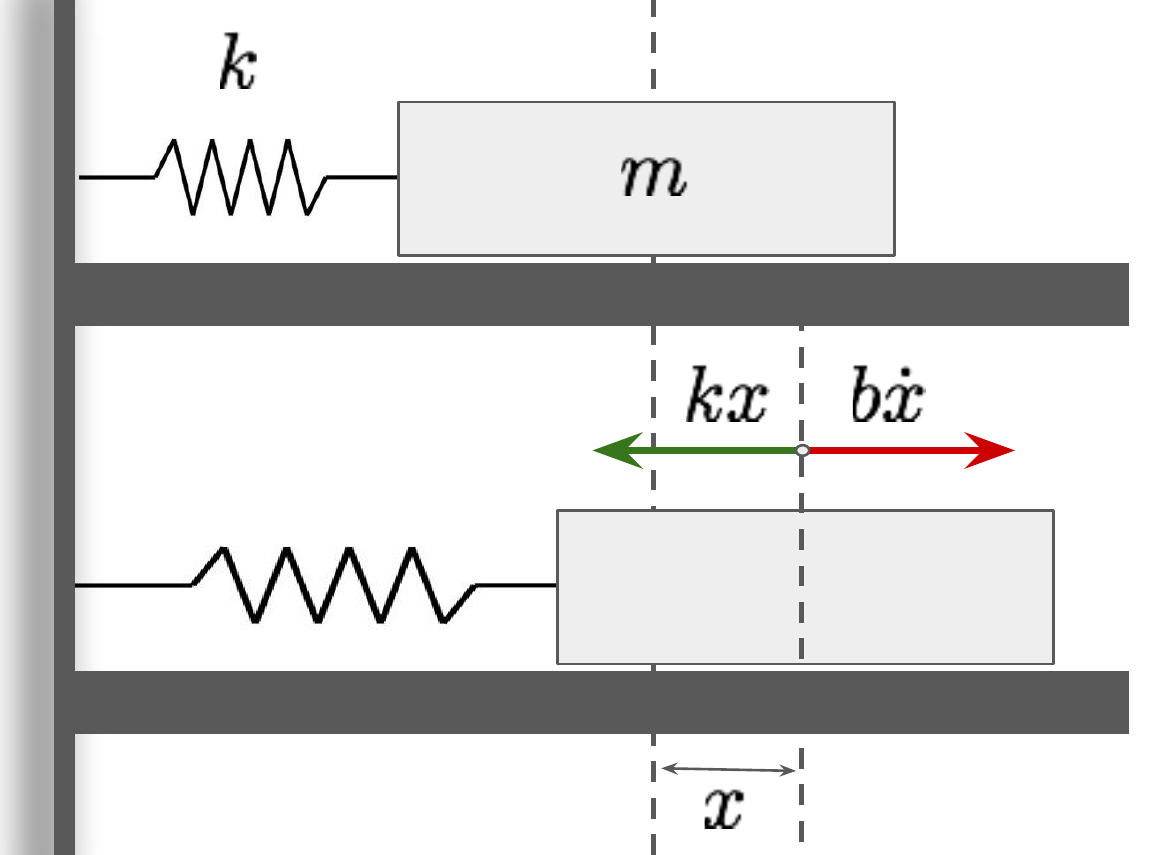}
	\caption{Elastic pendulum}
	\label{eq:setup}
\end{figure}

The damped harmonic oscillator is a simple dynamical system subject to dissipation (see Figure \ref{eq:setup}), described by the differential equation:

\begin{equation}
	\frac{d^2 x}{dt^2} = -\frac{b}{m}\frac{d x}{dt} - \frac{k}{m} x.
	\label{eq:dissipation2}
\end{equation} 
We replace the derivatives with their central difference approximation and get:

\begin{equation}
	\bigg(\frac{x_{t+\Delta}-2x_t + x_{t-\Delta}}{\Delta^2} \bigg) \approx \frac{-b}{m} \bigg(\frac{x_t - x_{t-\Delta}}{\Delta} 		\bigg) - \frac{k}{m} x_t
	\label{eq:dissipation3}
\end{equation} 
\begin{equation}
	x_{t+\Delta} \approx \frac{-b\Delta}{m}(x_t-x_{t-\Delta})-\frac{k\Delta^2}{m} x_t+2x_t-x_{t-\Delta}
	\label{eq:dissipation4}
\end{equation} 
\subsubsection{Canonical weights}
Three physical constants govern Equation \ref{eq:dissipation4}: the mass $m$, the spring constant $k$ and the damping coefficient $b$. From now on we will refer to them as canonical weights, to be learned by the solver network:
\begin{equation}
	\mathcal{W} = [m,b,k].
\end{equation}
We chose the canonical weights of the network to reflect the proper parametrization of our ODE for explanatory reasons only. As we will reveal in the result section, this choice allows us to demonstrate the capabilities of OscillatorNet in an intuitive manner. However if we want the learned values to be expressed in their proper SI units we have to initialize them with values of the correct order of magnitude. In Section \ref{sec:combined_weights} we generalize these results by training more general weights only dependent on the sampling frequency of the input time-series.\\

For two input vectors $\mathbf{x}$ and $\mathbf{x}^{t-\Delta}$ we can express a finite difference step of the network in terms of the canonical weights as:
\[  
  \left[
    \begin{array}{cccc}
      \frac{k\Delta^2}{m}& \frac{-b\Delta}{m}
    \end{array}
  \right]\cdot
  \left[
    \begin{array}{c}
      \mathbf{x}^t\\
      \mathbf{x}^t-\mathbf{x}^{t-\Delta}
    \end{array}
  \right]
\]
\subsubsection{Experiments}
In our experiment we input 60 consecutive real-space trajectory points of a damped harmonic oscillator with mass $m = 1 \si{kg}$, damping coefficient $b=0.8 \si{kg/s}$ and spring constant $k=40 \si{kg/s^2}$  in a 1-layer OscillatorNet (kernel size = 1, channels = 2, i.e. number of free parameters = 3). We can now carry out a single finite difference step while the parameters of the differential equation are automatically learned from the data. If we choose to forecast multiple points (free forecast), each prediction is fed back into the layer, before the network is re-trained to generate the next forecast.\\ 

\begin{figure}[htb!]
	\centering
	\includegraphics[width=1\linewidth]{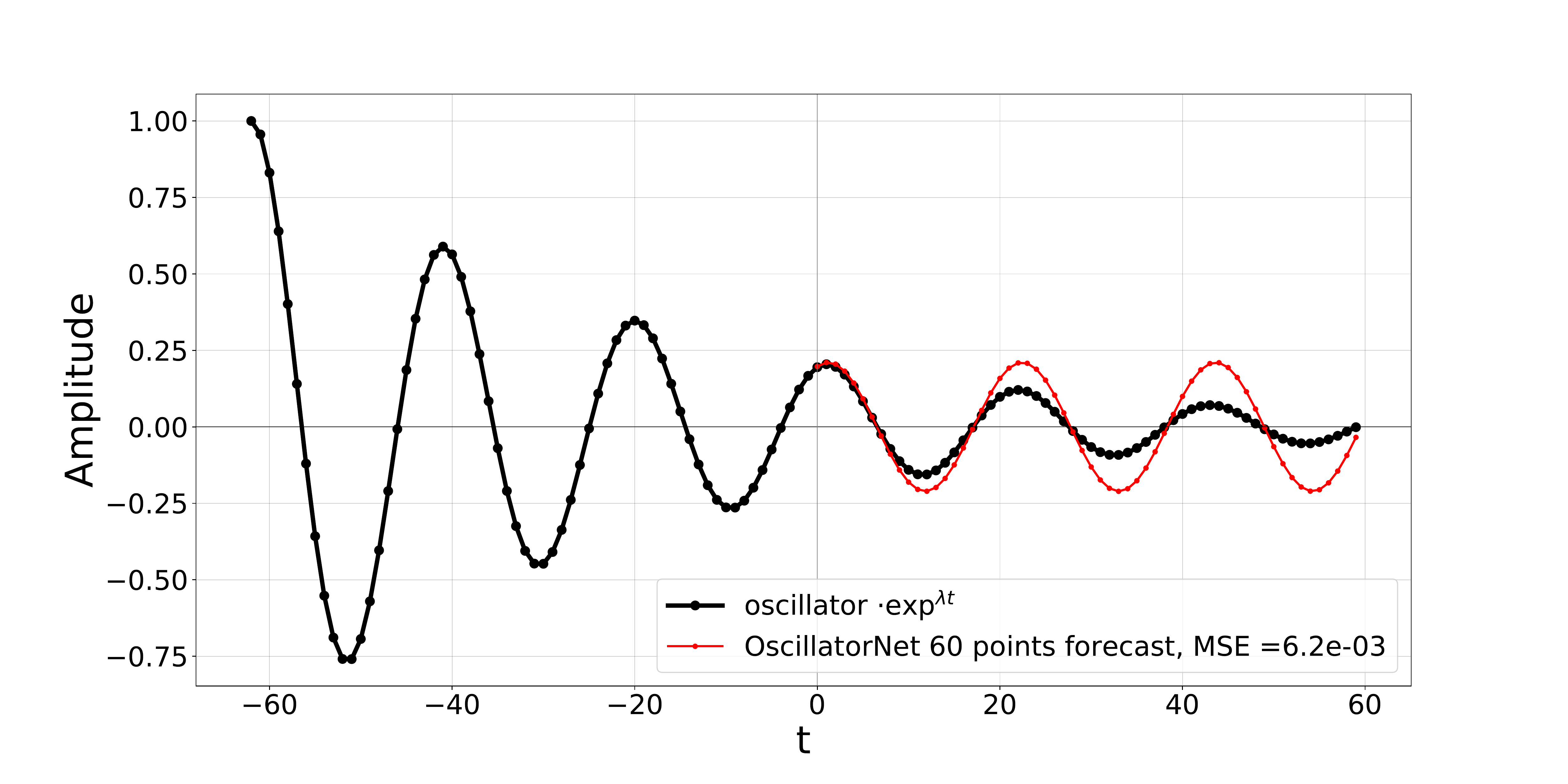}
	\centering
	\includegraphics[width=1\linewidth]{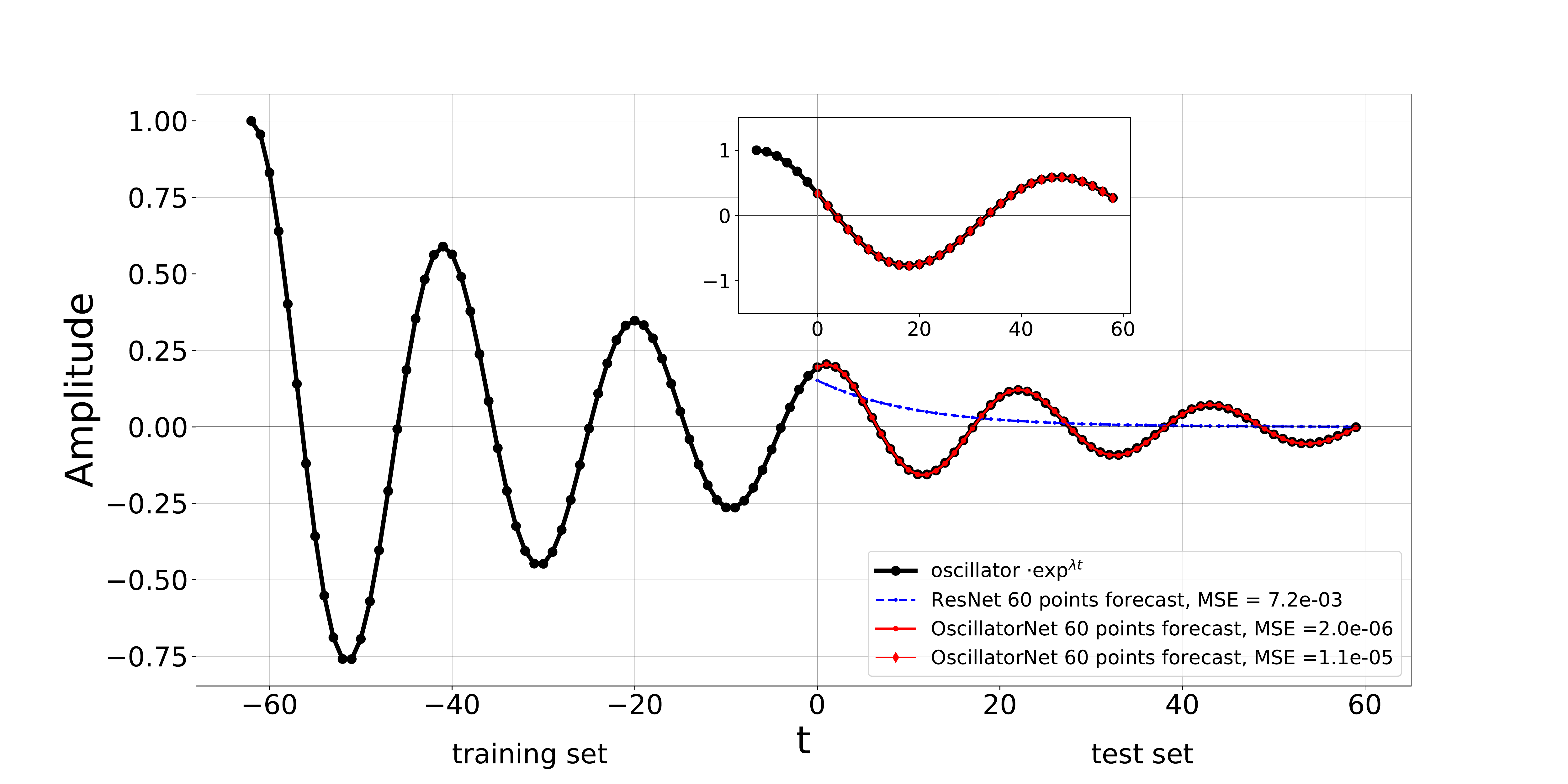}
	\caption{\emph{Top panel:} Output from OscillatorNet without doubling the number of filters, i.e. the network cannot learn the dissipative term oft he governing equation. \emph{Botton panel:} The network has enough free parameters to 	learn the full equation $\ddot{x} = f(x,\dot{x})$. The blue dots represent the output of 1 layer ResNet. \emph{Inset:} Forecast if the network is trained on less than $1/4 \pi$ of the oscillator signal.}
	\label{fig:single_oscillator}
\end{figure}

Figure \ref{fig:single_oscillator} shows the results of a free forecast of 60 real-space trajectory points, i.e. except for the training set, no data of the time-series is used to train the network. The top panel shows the networks output with only one  filter. We learn the correct frequency but the amplitude deviates increasingly as a function of time, due to the fact that we cannot learn a second inhomogeneous term as function of the oscillators momentum and therefor the total energy is conserved.

The lower panel of Figure \ref{fig:single_oscillator} shows the forecast of OscillatorNet with two filters. In comparison we also show the result of a single ResNet layer (blue dots). \\
The fact that we can forecast multiple time-steps in a stable fashion, without exposing the network to more ground-truth data, indicates that we indeed learn the differential equation. To confirm this claim and to show that we accurately learn the governing physics of the system, we extract the trained canonical weights. A comparison of the true values versus the canonical weights learned by OscillatorNet can be found in Table \ref{tab:single_oscillator_res}.\\
\begin{table}[htb!]
\centering
	\begin{tabular}{ |c||c|c|c| }
 	\hline
 	\multicolumn{4}{|c|}{\textbf{ Learned Parameters of a single oscillator in SI units}} \\
 	\hline
 	 & True value &OscillatorNet $\mathcal{W}$&Init.$\mathcal{W}$\\
	 \hline
	 \hline
	 Mass [\si{kg} ]  & 2  & 2.058& 1\\
	 \hline
	 Damping [\si{kg/s}]&   1.5 & 1.487&1   \\
	\hline
	 Spring constant [\si{kg/s^2}]& 40  & 40.249& 15\\
	\hline
	\end{tabular}
	\caption{}
	\label{tab:single_oscillator_res}
\end{table}
An impressive feature of the network is it's ability to approximatively learn the correct parameters even if it is only trained on a very small sample of the time-series. The inset in Figure \ref{fig:single_oscillator} shows the output of the network if trained on less than a quarter period of the damped oscillator signal. The learned parameters can be found in Table \ref{tab:single_oscillator_res_inset}.
%\subsection{Signal generator}
%TBD
%To have a high flexibility in testing our network, we use a signal generator. where we can build a system of coupled harmonic oscillators 
%If we are purely interested in a stable forecast of the system, we don't have to express the trainable %parameters in terms of canonical weights, instead the ratios of the canonical weights are learned
%(frequency $1/s$ and it's derivative $1/s^2$) (needs some works and should be shifted to the canonical %weights section).  
\begin{table}[htb!]
\centering
	\begin{tabular}{ |c||c|c|c| }
 	\hline
 	\multicolumn{4}{|c|}{\textbf{ Training on less than} $\mathbf{1/4} \mathbf{\pi}$} \\
 	\hline
 	 & True value &OscillatorNet&Init.$\mathcal{W}$\\
	 \hline
	 \hline
	 Mass [\si{kg} ]  & 2  & 1.918& 1\\
	 \hline
	 Damping [\si{kg/s}]&   1.5 & 1.343&1   \\
	\hline
	 Spring constant [\si{kg/s^2}]& 40  & 38.80& 15\\
	\hline
	\end{tabular}
	\caption{}
	\label{tab:single_oscillator_res_inset}
\end{table}

\subsection{Coupled harmonic oscillators}

\begin{figure}[b!]
\centering
\includegraphics[width=0.7\linewidth]{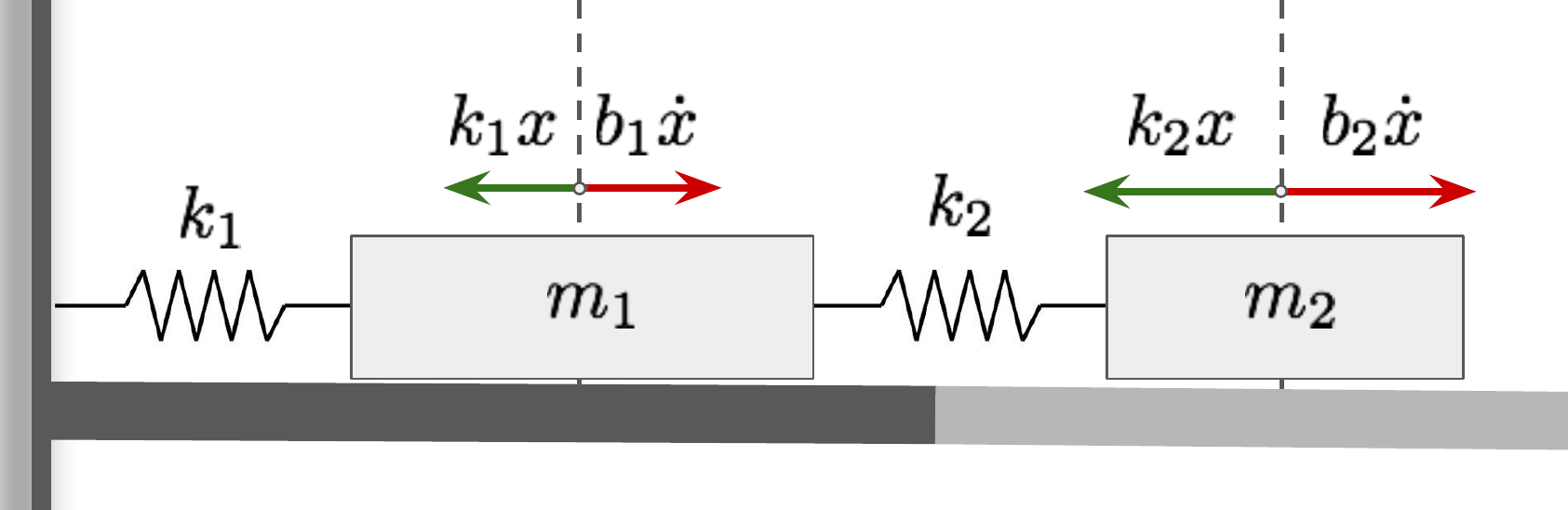}
\caption{Damped coupled harmonic oscillator}
\label{fig:setup2}
\end{figure}

\begin{figure}[htb!]
	\centering
	\includegraphics[width=1\linewidth]{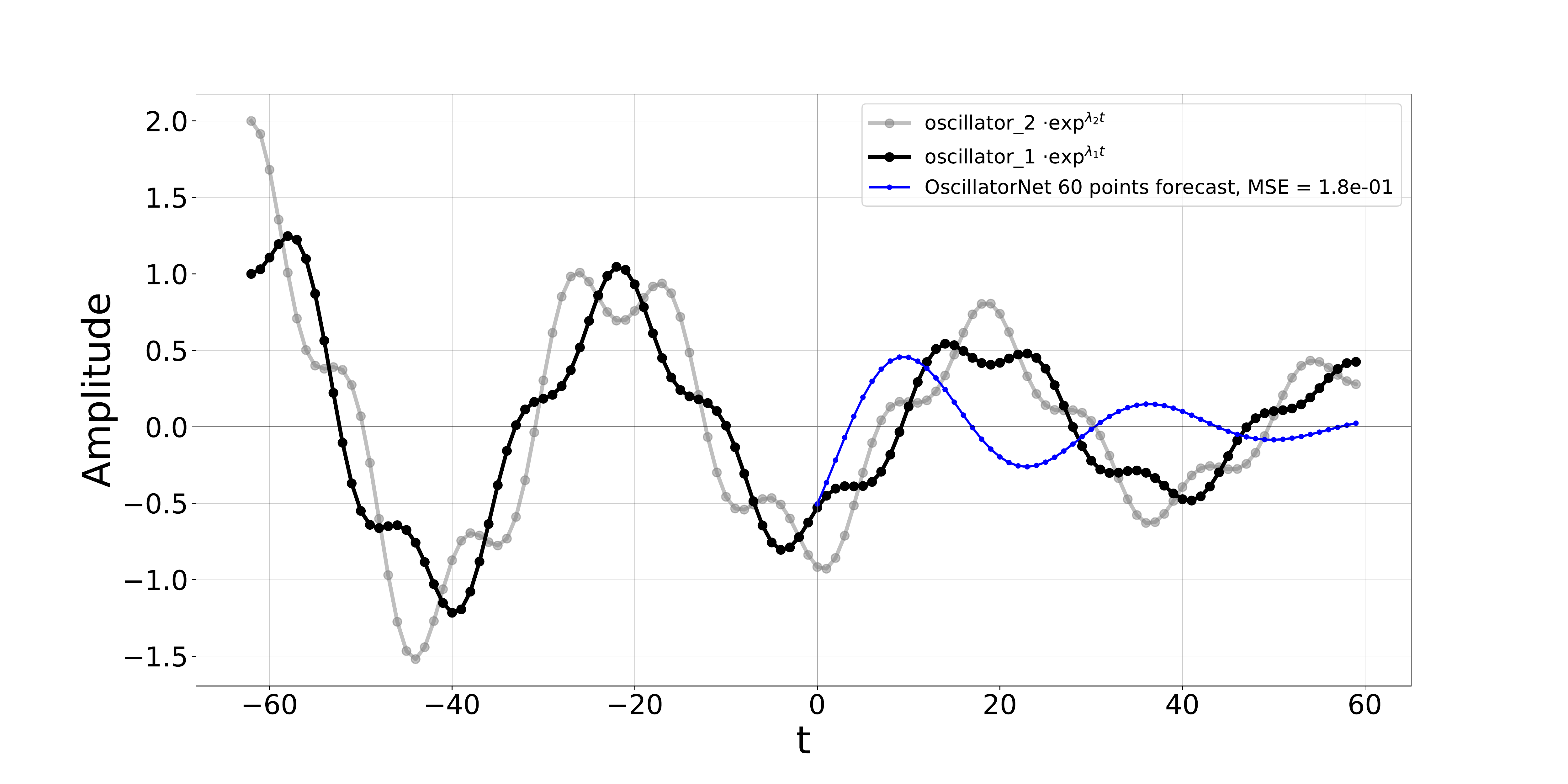}
	\includegraphics[width=1\linewidth]{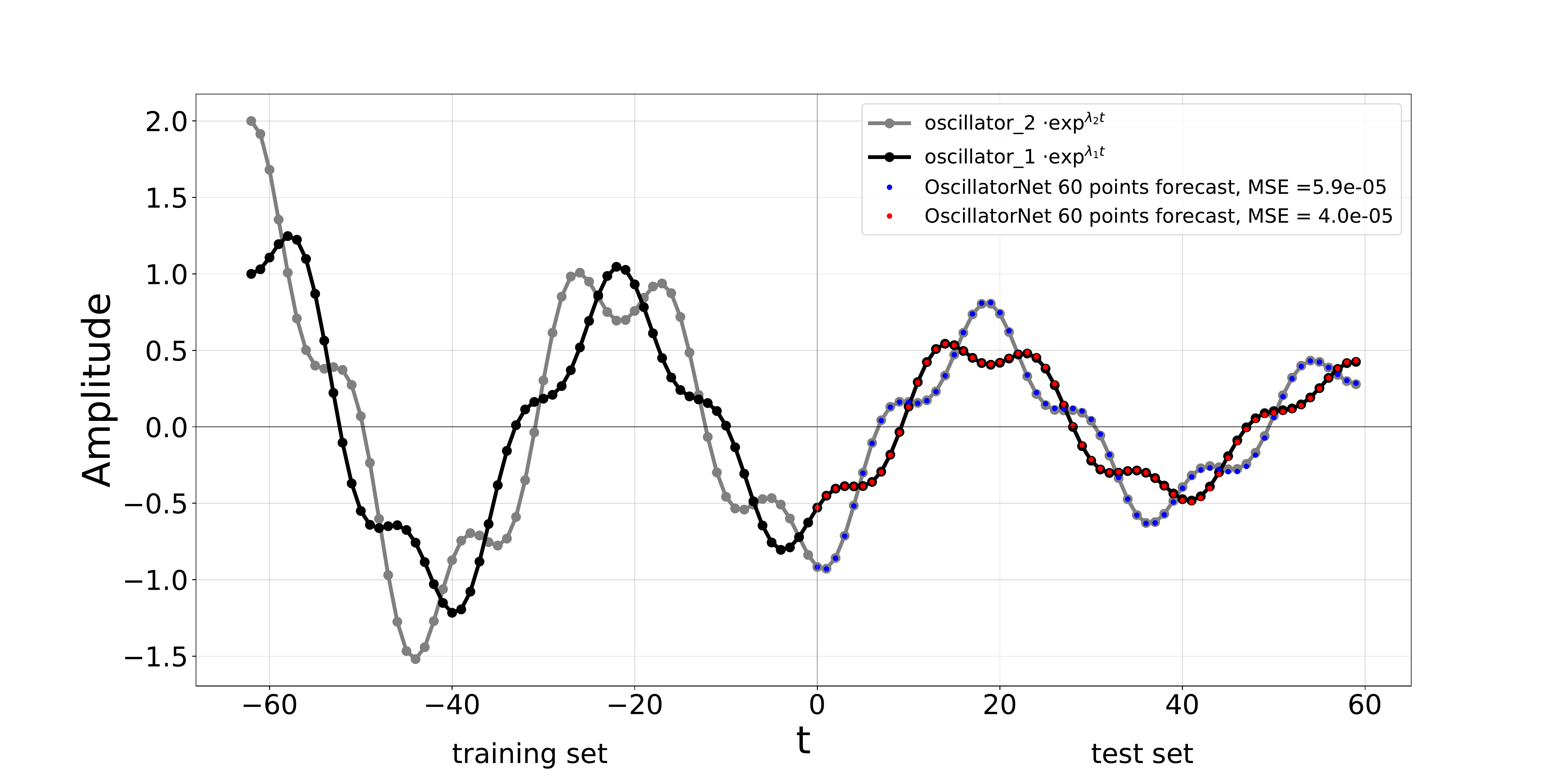}
	\caption{\emph{Top panel:} 60 points free forecast of one trajectory of a coupled harmonic oscillator after the network is trained on only on the trajectory of oscillator 1. \emph{Lower panel:} 60 points free forecast of the full oscillator system after training on 60 points of each oscillator trajectory.}
	\label{fig:coupled_oscillator}
\end{figure}

OscillatorNet generalizes to arbitrary many coupled masses. Let's take a look at two coupled oscillators (Figure \ref{fig:setup2}), subject to different spring constants and damping coefficients, described by the system of ODEs:

\begin{eqnarray}
	\frac{d^2{x}_1}{dt^2} &=& \frac{-b_1}{m_1}\frac{d{x}_1}{dt} - \frac{k_1}{m_1} x_1 +(x_2 - x_1)\frac{k_2}{m_1} \nonumber\\
	\frac{d^2{x}_2}{dt^2}  &=& \frac{-b_2}{m_2}\frac{d{x}_2}{dt} -(x_2 - x_1)\frac{k_2}{m_2} 
	\label{eq:system2}
\end{eqnarray}
Replacing the derivatives with the respective central difference approximations and solving for $\mathbf{x}^{t+\Delta}$ becomes:
\begin{eqnarray}
	x_1^{t+\Delta} &\approx&\frac{-b_1\Delta}{m_1}(x_1^t-x_1^{t-\Delta})-\frac{\Delta^2}{m_1}(k_1+k_2)x_1^t +\frac{k_2\Delta^2}{m_1}x_2^t +2x_1^t -x_1^{t-\Delta} \nonumber\\
	x_2^{t+\Delta} &\approx&\frac{-b_2\Delta}{m_2}(x_2^t-x_2^{t-\Delta})-\frac{k_2\Delta^2}{m_2}(x_1^t +x_2^t ) +2x_2^t -x_2^{t-\Delta}	
\end{eqnarray}
The canonical weights of this problem are:
\begin{equation}
	\mathcal{W} =[m_1,m_2,b_1,b_2,k_1,k_2]      
\end{equation}
For two input time-series $[\mathbf{x}_1, \mathbf{x}_1^{t-\Delta}$ ]and $[\mathbf{x}_2,\mathbf{x}_2^{t-\Delta}] $ we can express a finite difference step of the network in terms of the canonical weights as:
\[  
  \left[
    \begin{array}{cccc}
      \frac{-\Delta^2}{m_1}(k_1+k_2)& \frac{\Delta^2}{m_1}k_2& \frac{-\Delta b_1}{m1}&0\\
      \frac{\Delta^2}{m_2}k_2& \frac{-\Delta^2}{m_2}k_2&0& \frac{-\Delta b_2}{m2}
    \end{array}
  \right]\cdot
  \left[
    \begin{array}{c}
      \mathbf{x}_1^t\\
      \mathbf{x}_2^t\\
       \mathbf{x}_1^t-\mathbf{x}_1^{t-\Delta}\\
        \mathbf{x}_2^t-\mathbf{x}_2^{t-\Delta}
    \end{array}
  \right]
\]
We train OscillatorNet on 60 points of the signal by embedding each trajectory as a separate channel into the network. The network then carries out a single finite difference step for both time-series embeddings, while learning the parameters of the differential equations, before we feed the forecast back into the network and retrain to produce the next forecast. The lower panel in Figure \ref{fig:coupled_oscillator} shows the results of the free forecast of 60 points for each trajectory, achieved in this manner. The learned weights can be found in Table \ref{tab:coupled_oscillator_res}.

\begin{table}[htb!]
\centering
	\begin{tabular}{ |c||c|c|c| }
 	\hline
 	\multicolumn{4}{|c|}{\textbf{ Learned Parameters of a coupled oscillator in SI units}} \\
 	\hline
 	 & True value &OscillatorNet $\mathcal{W}$&Init.$\mathcal{W}$\\
	\hline
	\hline
	Masses  [\si{kg}]   & 1.5, 0.9 & 1.514, 0.906 &1, 1  \\
	\hline
	Dampings [\si{kg/s}]&   0.5 , 0.3 & 0.483 , 0.297  & 1, 1\\
	\hline
	Spring constants [\si{kg/s^2}]& 14, 35 & 14.013, 34.472& 15, 15\\
	\hline
	\end{tabular}
	\caption{}
	\label{tab:coupled_oscillator_res}
\end{table}

\section{Exploiting the relation between real-space trajectories of a coupled system} \label{exploit}
In a real-world application we might not be able to observe the complete configuration space of a coupled dynamical system. Let's consider again the toy-model consisting of two coupled masses depicted in Figure \ref{fig:setup2}. If only partial information about the system is available, i.e. just one time-series describing the trajectory of the first oscillator, a stable forecast is not possible (see top panel of Figure \ref{fig:coupled_oscillator}).
Let's go back to a system of coupled ODEs as in Equation \ref{eq:system2}. If the boundary conditions are fixed on one side, we can recover the position in time of a second coupled oscillator from the first.  If we solve the first equation for $x_2$ and replace the derivative with it's proper backward difference approximation, we arrive at :

\begin{equation}
	x_{2}^t = \frac{1}{k_2}\{\frac{m_1}{\Delta^2}(x_1^t - 2^{t-\Delta}x_1+x_1^{t-2\Delta}) +\frac{b_1}{\Delta}(x_1^t - x_1^{t-\Delta})+k_1x_1^t\}+x_1^t.
	%\label{eq:second_mass}
\end{equation}
If we have more than two coupled oscillators in the system of consideration, the $i^\text{th}$ oscillator position is given by:
\begin{eqnarray}
	x_{i}^t = \frac{1}{k_i}\{\frac{m_{i-1}}{\Delta^2}(x_{i-1}^t - 2^{t-\Delta}x_{i-1}+x_{i-1}^{t-2\Delta})		\label{eqn: many_mass} \\ \nonumber 
	 +\frac{b_{i-1}}{\Delta}(x_{i-1}^t - x_{i-1}^{t-\Delta}) \},\qquad 2< i\leq N 
\end{eqnarray}
Equation \ref{eqn: many_mass} can be realized with a convolutional network of the form:
\begin{equation}
	x_{2}^t =  \alpha \cdot \text{Conv}^3(x_1^t, [+1, -2, +1]) +
	 \beta \cdot \text{Conv}^2 (x_1^t, [+1, -1])+
	 \gamma\cdot x_1^t.
	\label{eq:stencil}
\end{equation}

\begin{figure}[htb!]
	\centering
	\includegraphics[width=1\linewidth]{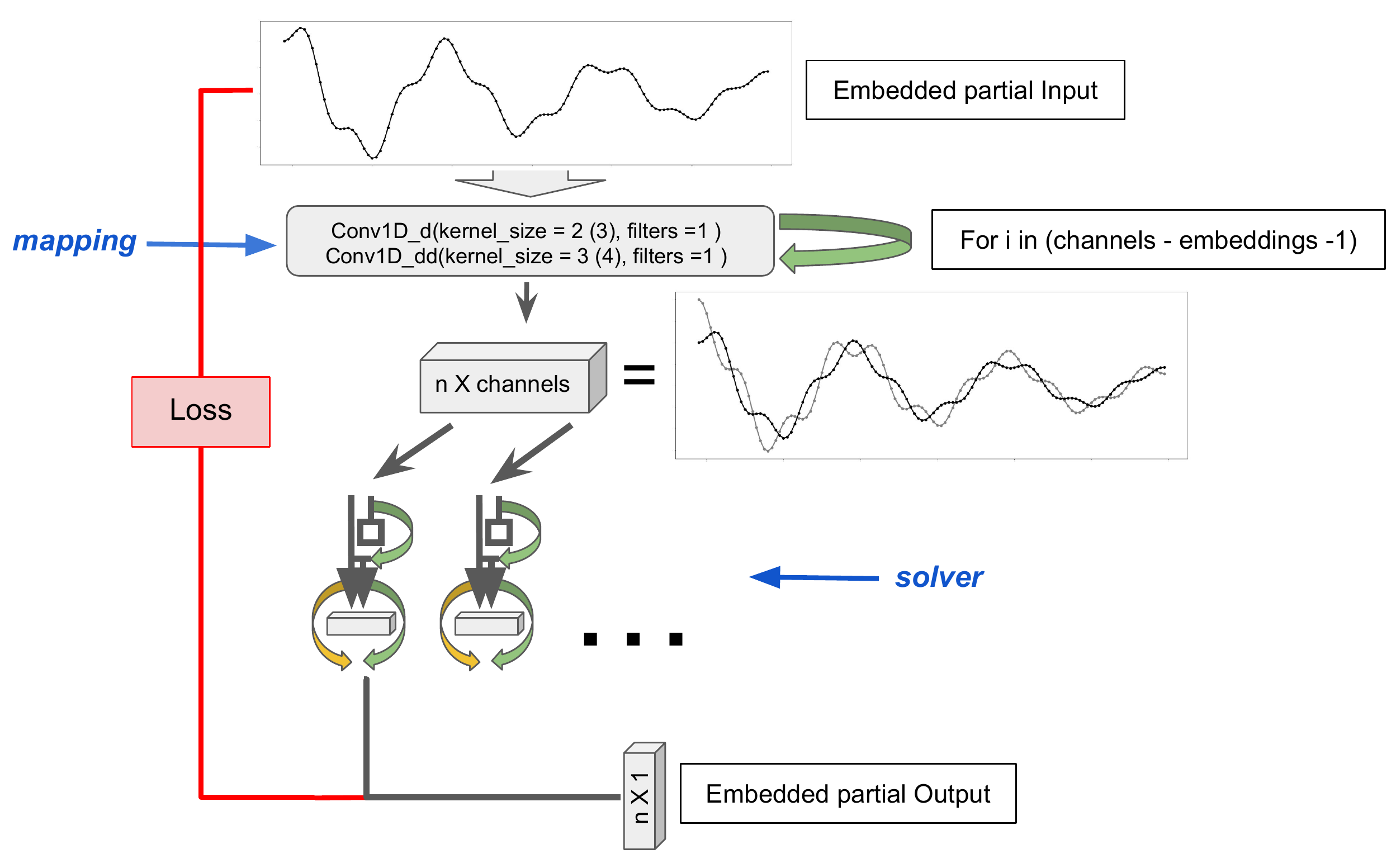}
	\caption{Schematic of the network - mapping \& solver }
	\label{fig:mapping_scheme}
\end{figure}

We refer to this part of the network as \textbf{mapping} (see Figure \ref{fig:mapping_scheme}), where we learn the \emph{relation} between $x_1$ and $x_2$ ($[x_1,x_2]$ and $x_3$ for 3 oscillators, etc.). The weights of the convolutional kernel in Equation \ref{eq:stencil} are fixed and represent a stencil on which we perform the numerical derivative. It is important that we use backward finite difference coefficients to respect the causal order of the time-series. The accuracy of this numerical derivative can be increased by changing its order. For example the second order stencil carries the following coefficients:

\begin{table}[htb!]
\centering
	\begin{tabular}{ |c|c|c|c| c|}
	\hline
	 grid position: & -3&-2&-1&0\\
	\hline
	\hline
	$1^{st}$ deriv.&   & 1/2  &-2 &3/2\\
	\hline
	$2^{nd}$ deriv.&-1 &4   &-5&2   \\
	\hline	
	\end{tabular}
	\label{tab:stencil}
	\caption{}
\end{table}

The 3 parameters $\alpha, \beta$ and $\gamma$ in Equation \ref{eq:stencil} however are trainable by an additional convolutional kernel that acts as a projection operator: 
\begin{eqnarray}
\alpha = \frac{m_1}{k_2\Delta^2},\quad
\beta =  \frac{b1}{k_2\Delta},\quad
\gamma = \frac{k_1+k_2}{k_2}
\end{eqnarray}
Since these parameters are also a combination of canonical parameters that form a subset $\mathcal{V}$ of $\mathcal{W}$, the canonical weights that where introduced in the  \textbf{solver} part of the network, it seems natural to share this subset over the whole network (solver and mapping). In other words, during training we update the weights of the mapping and the solver combined. However, since $\mathcal{V} \subset \mathcal{W}$, and we embed only one of the two trajectories in the network, when back-propagating the error not all canonical weights are updated but only the subset $\mathcal{V}$. This limits our capability in learning the physics that govern the system but not necessarily our ability to produce a stable forecast over a large horizon.

\subsection{Results}
It is only possible to share the parameters between mapping and solver, if we implement the projection according to Equation \ref{eq:stencil}. In other words, the choice of canonical weights dictates the size of the convolutional kernel that learns the linear combination of $\alpha, \beta$ and $\gamma$. For the canonical weights chosen in our examples the kernel is limited to size $1$. We can choose a wider kernel but have to cut the umbilical cord between solver and mapping. In our experiments we have tried both, a kernel of size $1$ that allows us to share $m_1, b_1, k_1$ and $k_2$, as well as a wider mapping kernel of size $25$, where we update $\mathcal{W}$ only on the solver side. 

\subsubsection{Canonical weights}

\begin{figure}[htb!]
	\centering
	\includegraphics[width=1\linewidth]{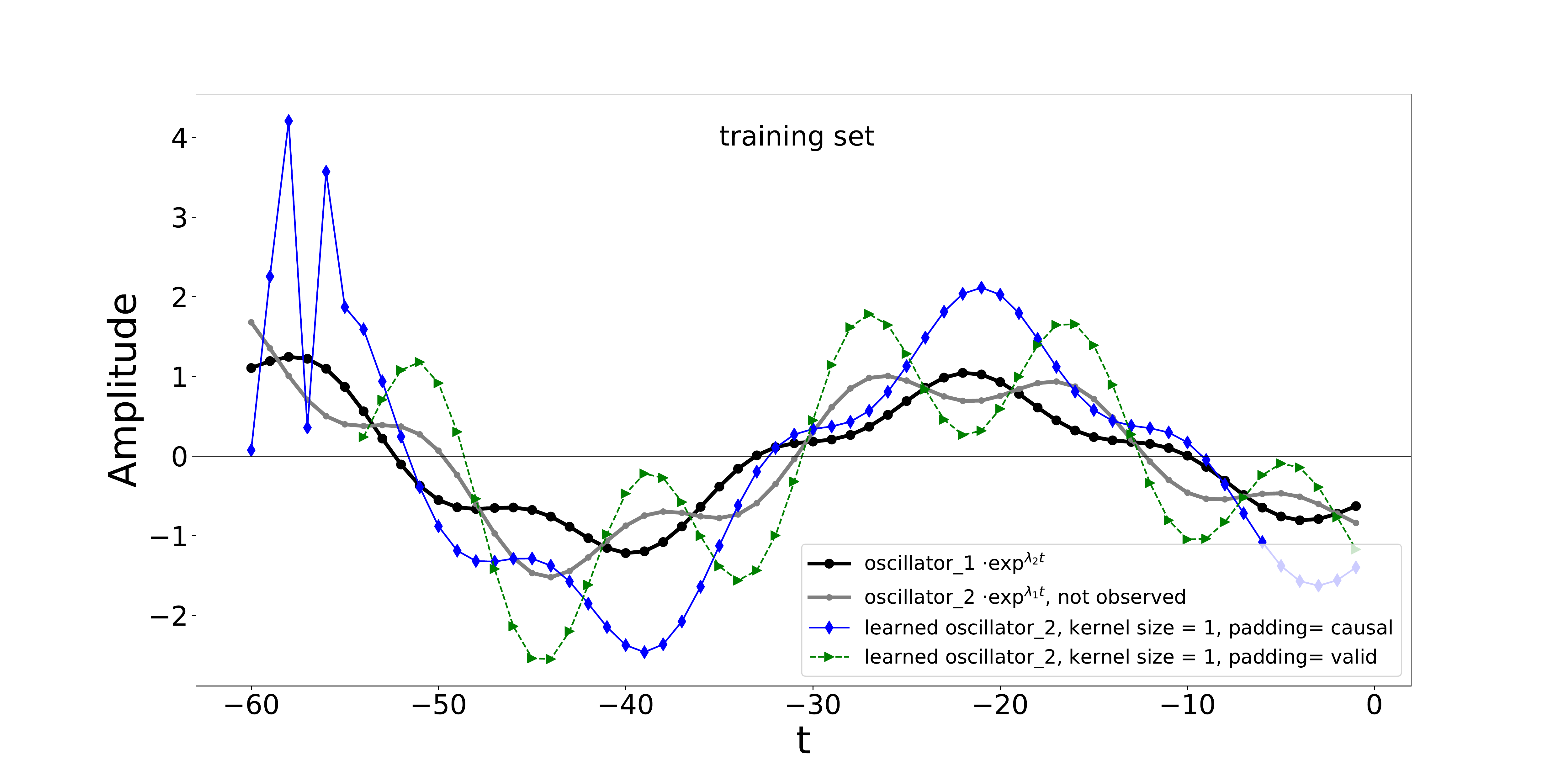}
	\includegraphics[width=1\linewidth]{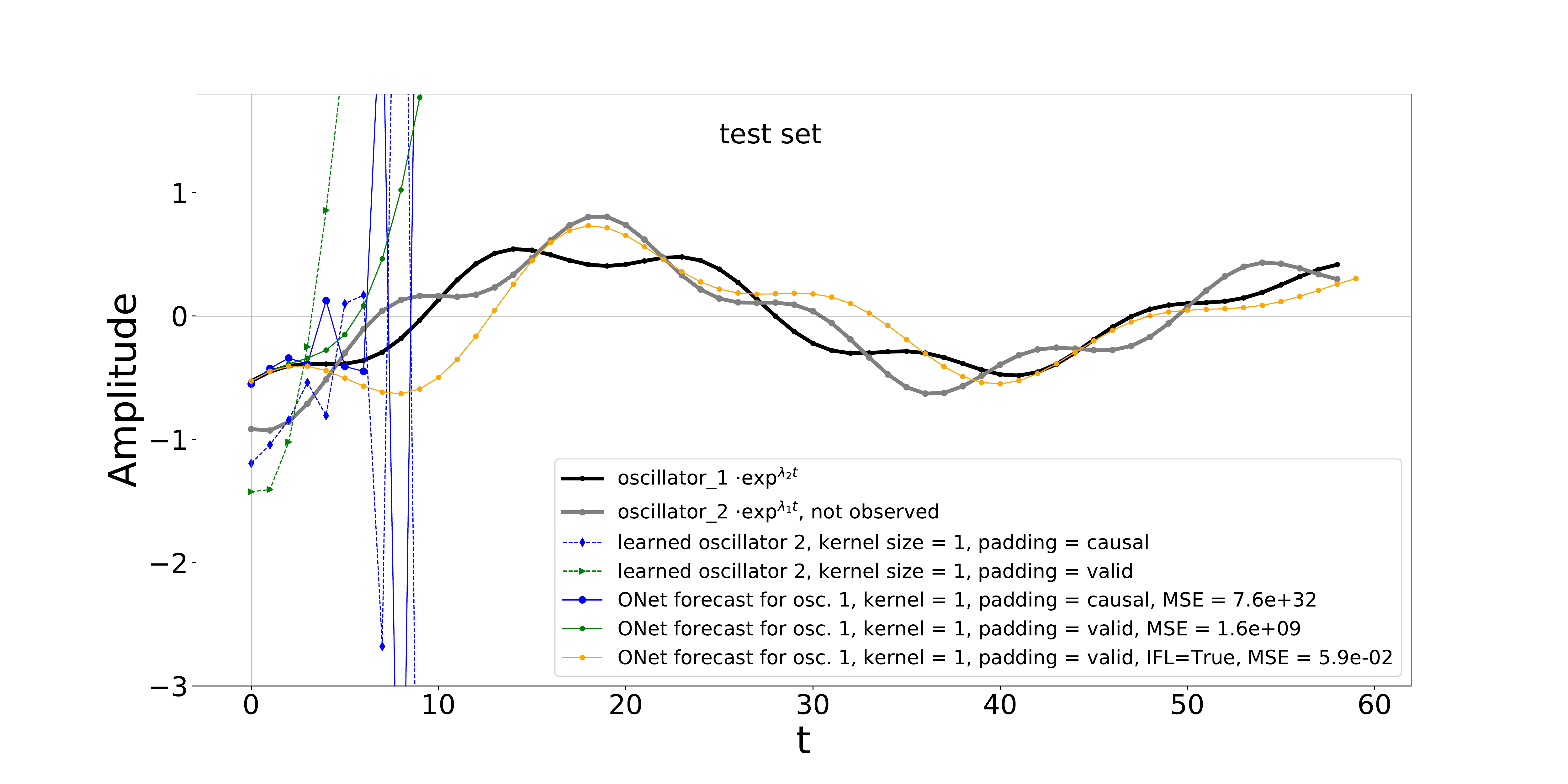}
	\caption{\emph{Top panel:} Network output of the mapping within the training set. \emph{Lower panel:} Mapping output for causal padding (blue diamonds) and valid padding (green triangles) and their respective free forecasts in the test set (for clarity only the first ten points are plotted). The orange line is the network output using valid padding but with the inner feedback loop activated.}
	\label{fig:mapping_size1}
\end{figure}
\begin{table}[]
\centering
	\begin{tabular}{ |c||c|c|c|c| }
 	\hline
 	\multicolumn{5}{|c|}{\textbf{OscillatorNet, learned parameters with partial input}} \\
	\multicolumn{5}{|c|}{\textbf{ padding = valid, mapping kernel = 1, stencil order = 5}} \\
 	\hline
 	 & True value &OscillatorNet $\mathcal{V}$&$\mathcal{V}$ with IFL &Init.$\mathcal{V}$\\
	\hline
	\hline
	$m_1$  [\si{kg}]   &1.5 & 1.535& 1.089&1  \\
	\hline
	$b_1$ [\si{kg/s}]&  0.5 & 0.216  & 0.090&1\\
	\hline
	$k_1$, $k_2$ [\si{kg/s^2}]& 14, 35 & 15.000, 15.000 & 15.000, 15.000&15, 15\\
	\hline
	\end{tabular}
	\caption{}
	\label{tab:mapping_size1_valid}
\end{table}
The first experiment shows the results of sharing the solver weights with the mapping, implemented according to Equation \ref{eq:stencil}, with a projection kernel of size one.  The blue diamonds show the mapping output if we set padding to causal, for the green triangles, padding has been set to valid. If we use zero padding during the training of the network, the numerical derivatives calculated on the mapping stencils are affected, as can be seen in the top panel of Figure \ref{fig:mapping_size1} between timestamp t=$-60$ and t=$-54$. If we set padding to valid, these points are truncated. What is more evident, is that even though if padding is set to valid, the mapping is not learning the correct trajectory for the second mass, it approximates a higher amplitude solution, i.e. all points where the trajectories of the two oscillators time-series intersect, are matched by the mapped trajectory. This result is not surprising since we update only a subset of all canonical weights when we back-propagate the error, the network is free to learn a solution that is numerically more stable, given it's partial degrees of freedom: $m_1$, $b_1$, $k_1$ and $k_2$, while $m_2$ and $b_2$ are fixed at initialization.\\

\begin{figure}[htb!]
	\centering
	\includegraphics[width=1\linewidth]{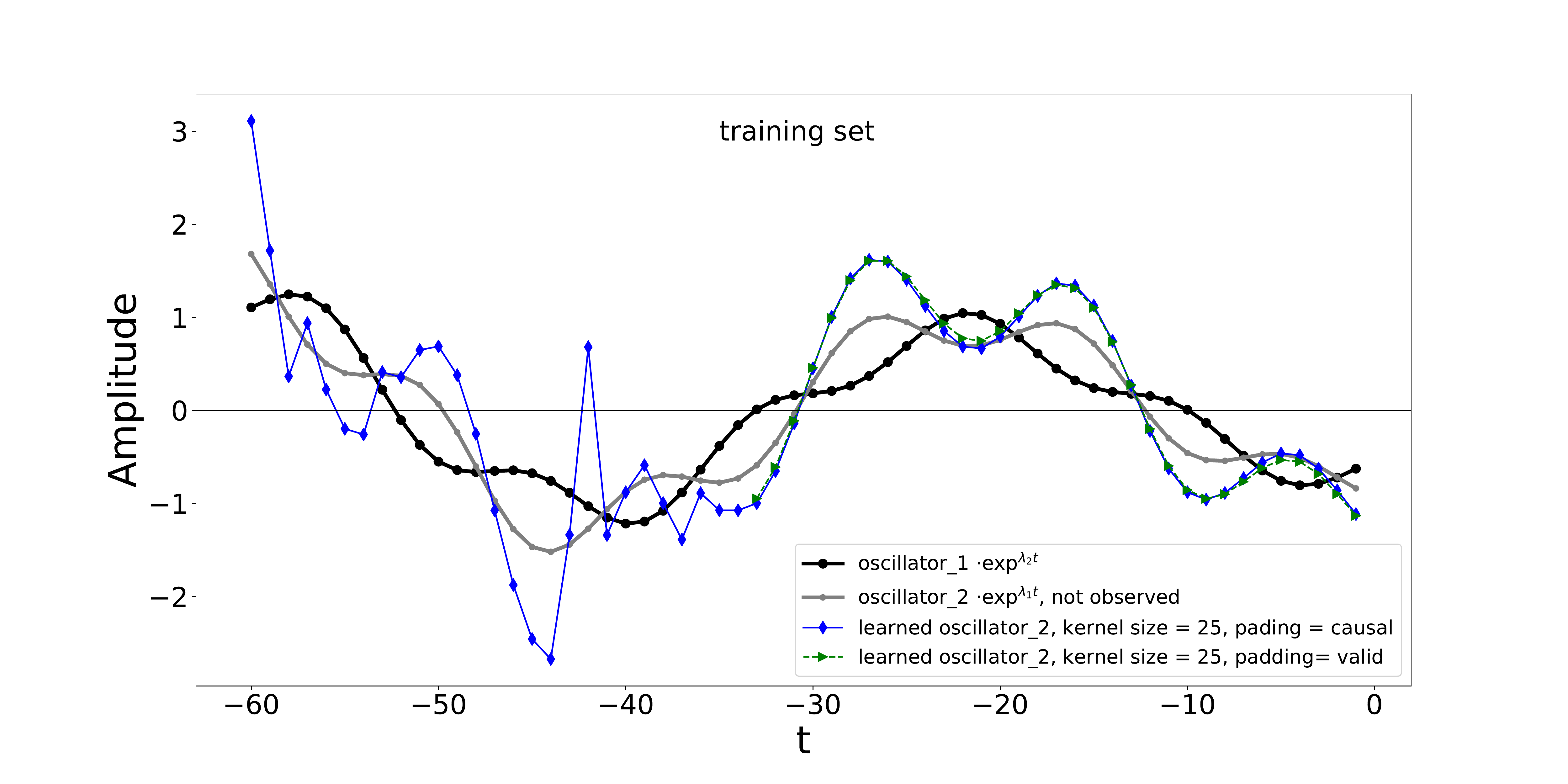}
	\includegraphics[width=1\linewidth]{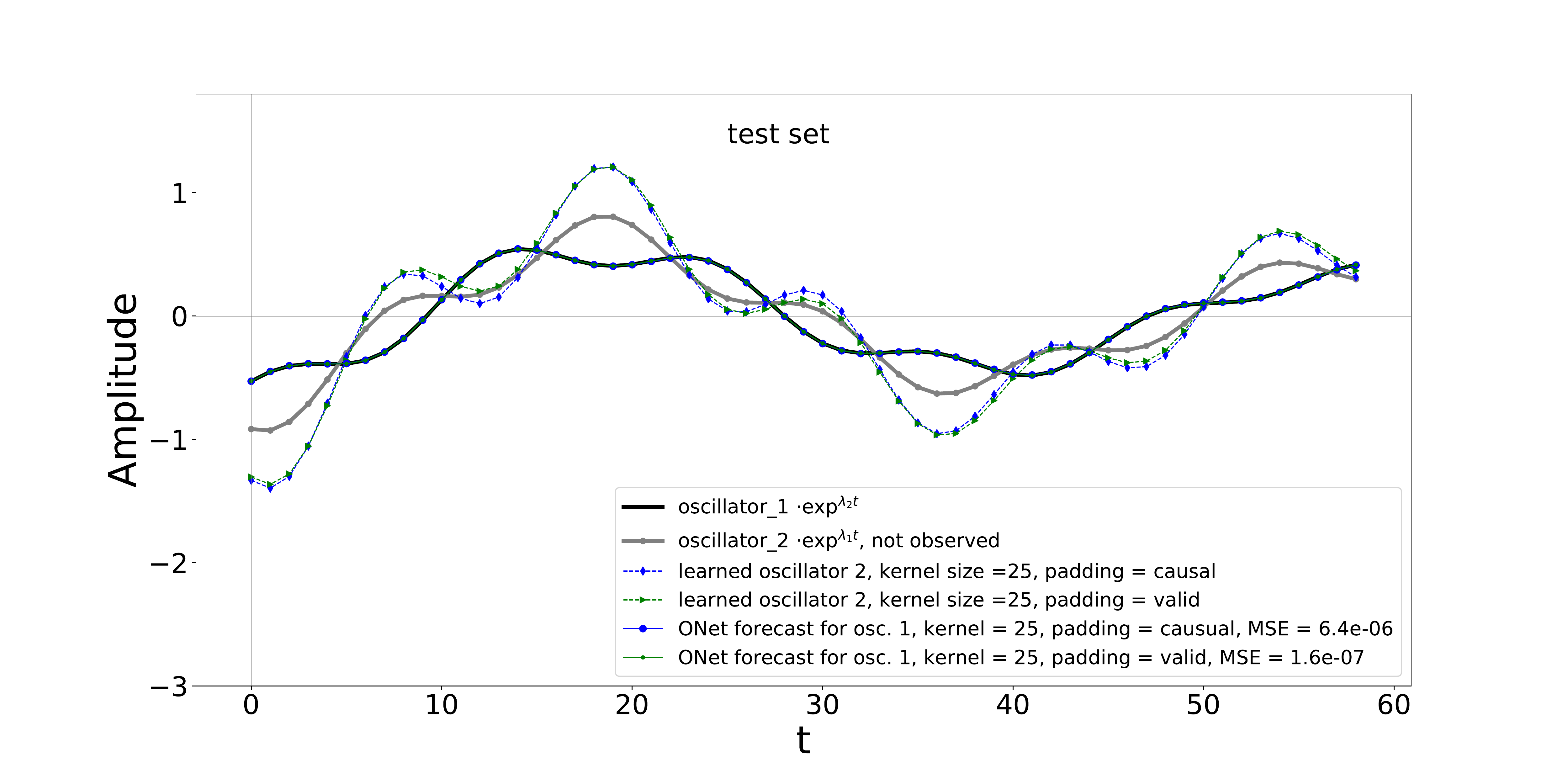}
	\caption{\emph{Top panel:} Blue diamonds show the quality of the mapping and the impact of zero padding, the green triangles show the mapping with padding set to valid. \emph{Botton panel:} Free forecast of oscillator 1 on basis of wide mapping kernel with padding set to valid (solid green line), and causal (solid blue line).}
	\label{fig:mapping_size25}
\end{figure}
The lower panel of Figure \ref{fig:mapping_size1} shows the out-of-sample free forecast based on the above training. Not surprisingly, the highest error is observed when padding is set to causal and therefore affecting the mapping kernel. But even the forecast with padding set to valid, proves to be only stable for the first 4 points before the error becomes too large. The forecast however can be stabilized, if we feed the mapped time-series back into the solver at each free-forecast step. The result of this inner feedback loop (IFL) is the orange curve in the lower panel of Figure\ref{fig:mapping_size1}. \\

If we widen the projection kernel to size 25, we are no longer able to share the weights between solver and mapping, but on the other hand we get more stable results in terms of the quality of the forecast. Figure \ref{fig:mapping_size25} shows the behaviour of the mapping network in the test and training set (blue diamonds for padding set to causal and green triangles for valid padding). If we take a look at the parameter learned (Table \ref{tab:mapping_25}), and compare it to Table \ref{tab:coupled_oscillator_res}, the results of embedding both trajectories into the network, we can see that we achieve a lower error when mapping to a higher frequency solution. 
%\\(The fact that we reach a more accurate forecast of oscillator 1, when learning the coupled oscillator 2 to be different than the ground-truth has to be more discussed) 
\begin{table}[htb!]
\centering
	\begin{tabular}{ |c||c|c|c| }
 	\hline
 	\multicolumn{4}{|c|}{\textbf{ OscillatorNet, learned parameters with partial input}}\\
	\multicolumn{4}{|c|}{\textbf{ padding = valid, mapping kernel = 25, stencil order = 5}} \\
 	\hline
 	 & True value &OscillatorNet $\mathcal{V}$&Init.$\mathcal{V}$\\
	\hline
	\hline
	Masses  [\si{kg}]   & 1.5 & 1.112 &1 \\
	\hline
	Dampings [\si{kg/s}]&   0.5  & 0.908  & 1\\
	\hline
	Spring constants [\si{kg/s^2}]& 14, 35 & 14.959, 14.827& 15, 15\\
	\hline
	\end{tabular}
	\caption{}
	\label{tab:mapping_25}
\end{table}
\begin{table}[htb!]
\centering
	\begin{tabular}{ |c||c|c|c| }
 	\hline
 	\multicolumn{4}{|c|}{\textbf{ OscillatorNet, learned parameters with partial input}}\\
	\multicolumn{4}{|c|}{\textbf{ padding = causal, mapping kernel = 25, stencil order = 5}} \\
 	\hline
 	 & True value &OscillatorNet $\mathcal{V}$&Init.$\mathcal{V}$\\
	\hline
	\hline
	Masses  [\si{kg}]   & 1.5 & 1.173 &1 \\
	\hline
	Dampings [\si{kg/s}]&   0.5  & 0.487  & 1\\
	\hline
	Spring constants [\si{kg/s^2}]& 14, 35 & 14.558, 15.091 & 15, 15\\
	\hline
	\end{tabular}
	\caption{}
	\label{tab:mapping_25_causal}
\end{table}
\subsubsection{Combined weights} \label{sec:combined_weights}
\begin{figure}[htb!]
	\centering
	\includegraphics[width=1\linewidth]{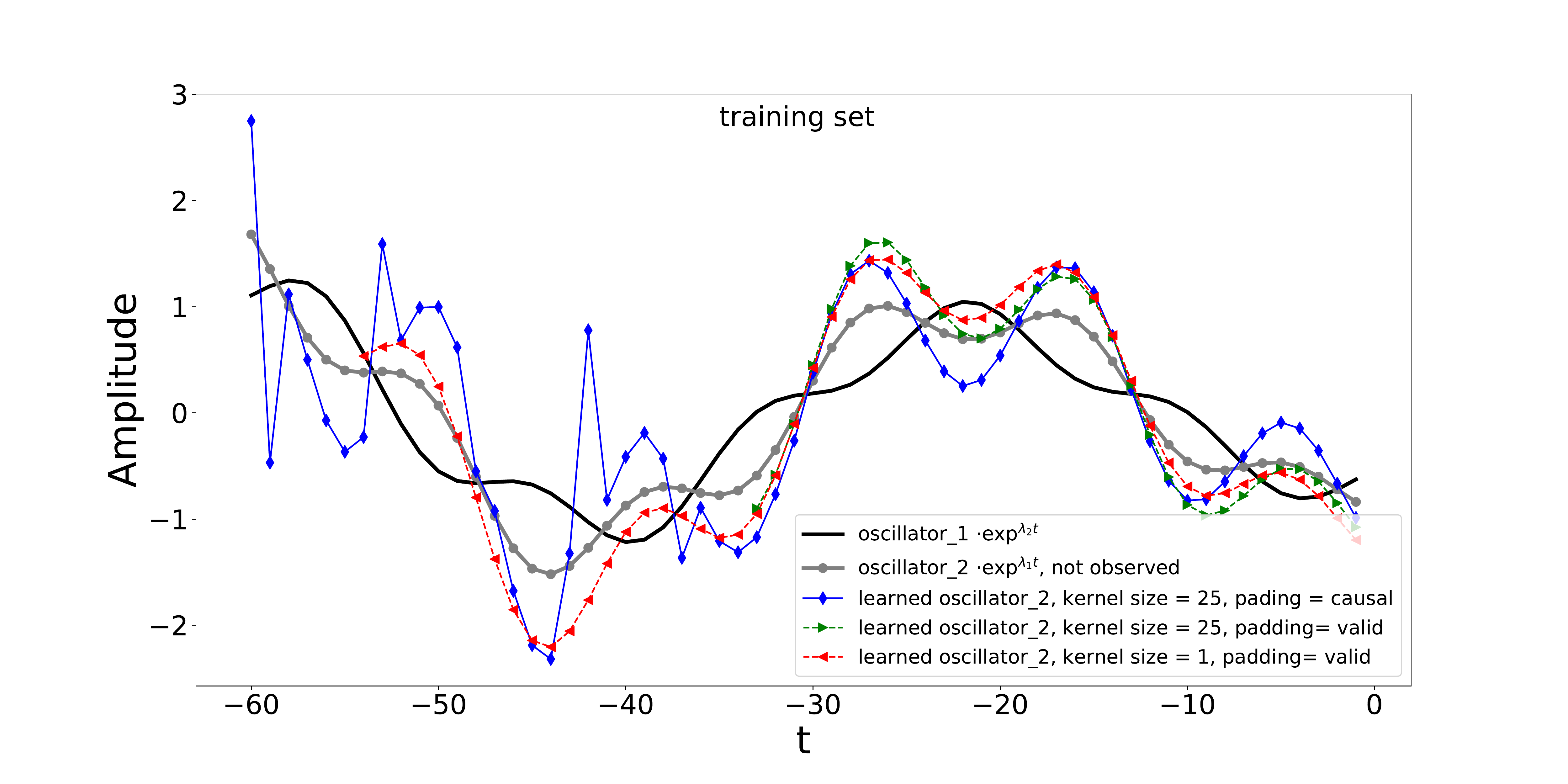}
	\includegraphics[width=1\linewidth]{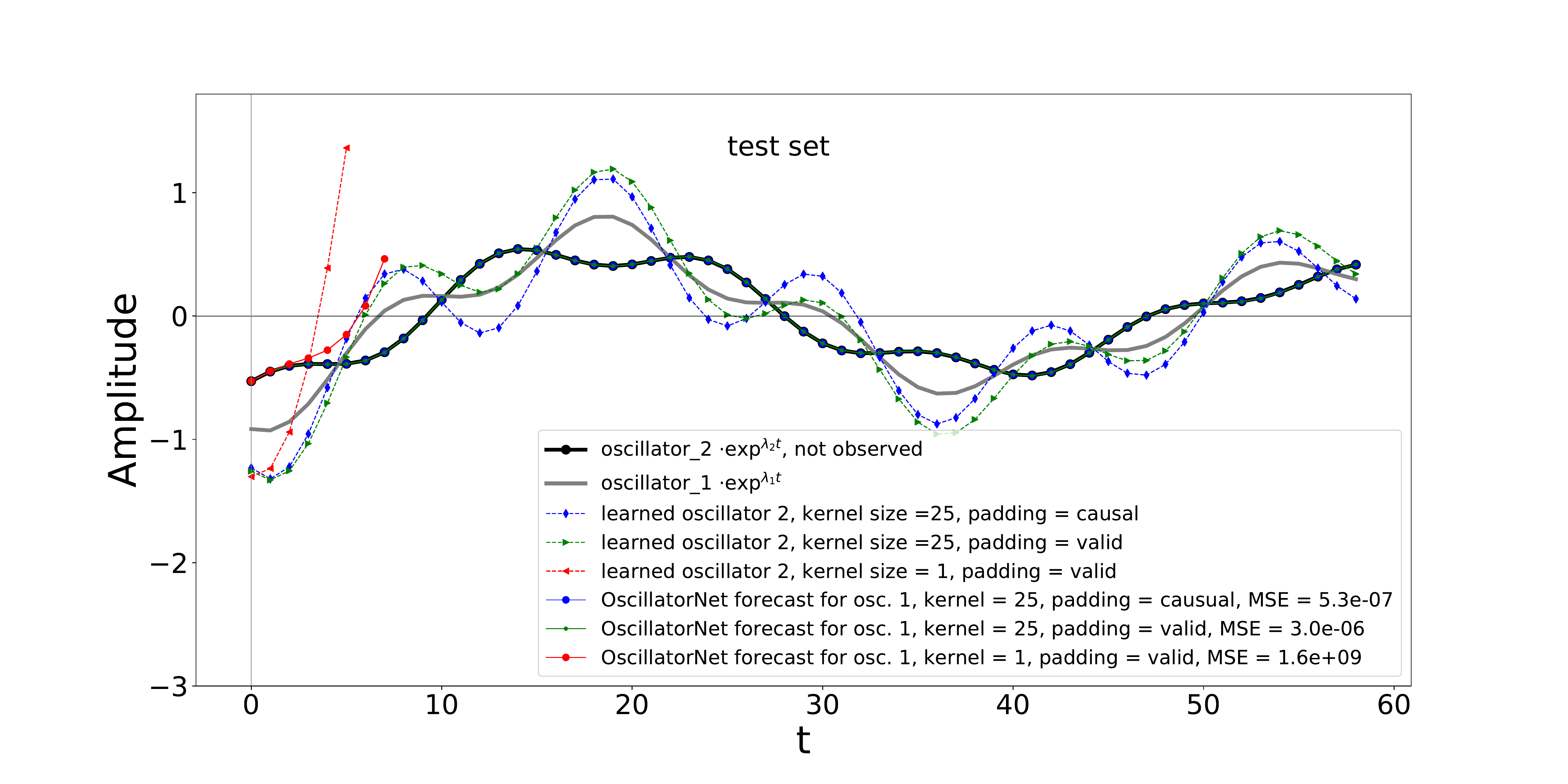}
	\caption{\emph{Top panel:} Mapping output during training for a wide mapping kernel and causal padding (blue diamonds), a wide mapping kernel and valid padding (green triangles) and for a mapping kernel of size 1 with valid padding (red triangles).\emph{Botton panel:} Test set showing the free forecast of different kernel sizes and padding. The mapping and forecast if kernel size is set to one (red triangles and dots) is only shown for the first 6 points due to large increase in error.}
	\label{fig:mapping_shared}
\end{figure}

When we look at the above results, we find that all examples where we use mapping, fail at approximating the second spring constant. The reason lies in the fact that the chosen weights make the problem numerically ill defined.  
In an attempt to make the network more general and ensure more numerical stability, we choose weights that only carry the units $[1/s]$ and $[1/s^2]$. In terms of the canonical weights, we can express the new set of parameters $\mathcal{U}$ as:
\begin{eqnarray}
\text{parameter}_a &=& \frac{\Delta^2 k_2}{m_1}\\
\text{parameter}_b &=& \frac{\Delta^2 k_2}{m_2}\\
\text{parameter}_c &=&  \frac{\Delta b_1}{m_1}\\
\text{parameter}_d &=& \frac{\Delta b_2}{m_2}\\
\text{parameter}_e &=&\frac{k_1+k_2}{k_2}
\end{eqnarray}
We initialize the network with the same values as before, for the combined weights, all results can be found in Figure \ref{fig:mapping_shared} and Tables \ref{tab:mapping_shared_causal},  \ref{tab:mapping_shared_valid} \& \ref{tab:mapping_shared_1}
\begin{table}[htb!]
\centering
	\begin{tabular}{ |c||c|c|c| }
 	\hline
 	\multicolumn{4}{|c|}{\textbf{OscillatorNet, learned parameters with partial input}}\\
	\multicolumn{4}{|c|}{\textbf{ padding = causal, mapping kernel = 25, stencil order = 5}} \\
 	\hline
 	 & True value &OscillatorNet $\mathcal{U}$&Init.$\mathcal{U}$\\
	\hline
	\hline
	parameter$_a$  $[\si{s}^{-2}]$   & 0.104 & 0.054 &0.066 \\
	\hline
	parameter$_b$ $[\si{s}^{-2}]$&   0.173  & 0.074  & 0.074\\
	\hline
	parameter$_c$ $[\si{s}^{-1}]$& 0.022 & 0.002& 0.066\\
	\hline
	parameter$_d$ $[\si{s}^{-1}]$&  0.022 & 0.022&  0.022\\
	\hline
	parameter$_e$ & 1.4 & 1.647 &2.0\\
	\hline
	\end{tabular}
	\caption{}
	\label{tab:mapping_shared_causal}
\end{table}

\begin{table}[htb!]
\centering
	\begin{tabular}{ |c||c|c|c| }
 	\hline
 	\multicolumn{4}{|c|}{\textbf{ OscillatorNet, learned parameters with partial input}}\\
	\multicolumn{4}{|c|}{\textbf{ padding = valid, mapping kernel = 25, stencil order = 5}} \\
 	\hline
 	 & True value &OscillatorNet $\mathcal{U}$&Init. $\mathcal{U}$\\
	\hline
	\hline
	parameter$_a$  $[\si{s}^{-2}]$   & 0.104 & 0.062 &0.066 \\
	\hline
	parameter$_b$ $[\si{s}^{-2}]$&   0.173  & 0.074  & 0.074\\
	\hline
	parameter$_c$ $[\si{s}^{-1}]$& 0.022 & 0.048& 0.066\\
	\hline
	parameter$_d$ $[\si{s}^{-1}]$&  0.022 & 0.022&  0.022\\
	\hline
	parameter$_e$ & 1.4 & 1.956 &2.0\\
	\hline
	\end{tabular}
	\caption{}
	\label{tab:mapping_shared_valid}
\end{table}
\begin{table}[htb!]
\centering
	\begin{tabular}{ |c||c|c|c| }
 	\hline
 	\multicolumn{4}{|c|}{\textbf{OscillatorNet, learned parameters with partial input}}\\
	\multicolumn{4}{|c|}{\textbf{ padding = valid, mapping kernel = 1, stencil order = 5}} \\
 	\hline
 	 & True value &OscillatorNet $\mathcal{U}$&Init. $\mathcal{U}$\\
	\hline
	\hline
	parameter$_a$  $[\si{s}^{-2}]$   & 0.104 & 0.065 &0.066 \\
	\hline
	parameter$_b$ $[\si{s}^{-2}]$&   0.173  & 0.074  & 0.074\\
	\hline
	parameter$_c$ $[\si{s}^{-1}]$& 0.022 & 0.009& 0.066\\
	\hline
	parameter$_d$ $[\si{s}^{-1}]$&  0.022 & 0.022&  0.022\\
	\hline
	parameter$_e$ & 1.4 & 1.999 &2.0\\
	\hline
	\end{tabular}
	\caption{}
	\label{tab:mapping_shared_1}
\end{table}

\section{Conclusion}
In this work we have introduced a new type of network, capable of solving the time-series patterns that form the solution of second order ODEs, subject to non-conservative forces. Instead of locally interpolating a time-series, the network introduced learns a prior representing the governing physical laws and is therefore capable to forecast the systems behaviour over a large time-horizon. Moreover we have shown that given the correct parametrization of the network we can approximate the governing equations to high accuracy. We have further investigated the possibility that, given only partial information about a coupled system we can train an internal mapping to retrieve a stable forecast of this partial system over many points.

\newpage

\bibliographystyle{ieeetr}
\bibliography{oscnet.bib}

\end{document}